%% file: main.tex
\documentclass{article}

\usepackage[final]{neurips_2021_ml4ad}

\usepackage[utf8]{inputenc} 
\usepackage[T1]{fontenc}    
\usepackage{hyperref}       
\usepackage{url}            
\usepackage{booktabs}       
\usepackage{amsfonts}       
\usepackage{nicefrac}       
\usepackage{microtype}      
\usepackage{xcolor}         
\usepackage{graphicx,dsfont,amssymb,booktabs,amsmath,subfigure}

\title{Switching Recurrent Kalman Networks}

\author{%
  Giao Nguyen-Quynh\\
  Karlsruhe Institute of Technology\\
  \texttt{uikon@student.kit.edu} \\
   \And
   Philipp Becker \\
    Karlsruhe Institute of Technology \\
   \texttt{	philipp.becker@kit.edu} \\
     \And
   Chen	Qiu \\
    Bosch Center for Artificial Intelligence \\
   \texttt{chen.qiu@de.bosch.com} \\
     \And
  Maja	Rudolph \\
    Bosch Center for Artificial Intelligence \\
   \texttt{maja.rudolph@us.bosch.com} \\
     \And
   Gerhard Neumann\\
    Karlsruhe Institute of Technology \\
   \texttt{gerhard.neumann@kit.edu} \\
}

\begin{document}

\maketitle

\begin{abstract}
 Forecasting driving behavior or other sensor measurements is an essential component of autonomous driving systems. Often real-world multivariate time series data is hard to model because the underlying dynamics are nonlinear and the observations are noisy. In addition, driving data can often be {\em multimodal} in distribution, meaning that there are distinct predictions that are likely, but averaging can hurt model performance. To address this, we propose the Switching Recurrent Kalman Network (SRKN) for efficient inference and prediction on nonlinear and multimodal time-series data. The model switches among several Kalman filters that model different aspects of the dynamics in a factorized latent state. We empirically test the resulting scalable and interpretable deep state-space model on toy data sets and real driving data from taxis in Porto. In all cases, the model can capture the multimodal nature of the dynamics in the data.
\end{abstract}

\section{Introduction}
Predicting the trajectory of a vehicle is a key competence of future autonomous driving. Future trajectory prediction refers to the estimation of the future state of some agents, given their past measurements. This ability is critical for autonomous vehicles to plan safe future navigations and avoid possible risks. Forecasting is a challenging task as there is an inherent ambiguity and uncertainty in predicting future trajectories. For example, at a given time instant of a traffic scene, there are several goals that a driver could have, and there are several plausible paths to reach each goal. Those goals are often not observable from the outside, making the future non-deterministic and multimodal at the same time. Averaging the dynamics is insufficient and in many cases physically impossible. Consider the scenario where there is an obstacle in the lane that a car is driving in. To avoid the obstacle, the car can change to the left lane or the right lane. Averaging these two possible maneuvers will lead the car to crash straight into the obstacle. The autonomous agents must be aware of these multiple possibilities to safely navigate through urban areas.

A common approach for modeling time series data is state-space models. They rely on latent states whose transition dynamics determine the system's behavior and are related to the measurements through a noisy observation process.
The Kalman filter \citep{kalman1960new} is the most widely used state-space model. It is the optimal solution for inferring linear Gaussian systems.
However, real-world time series data are often nonlinear, and the data generation process is unknown. Unfortunately, posterior inference in nonlinear non-Gaussian systems is generally intractable. There have been several efforts in the deep learning community to overcome the nonlinearity and system identification issue. Two common approaches are either to use approximations to make nonlinear systems tractable or to introduce stochasticity into recurrent neural networks \citep{schmidhuber1997long, chung2014empirical}. 

The Recurrent Kalman Network \citep{becker2019recurrent} (RKN)  is an efficient probabilistic recurrent neural network architecture that employs Kalman updates to infer the system state. In general, RKNs follow the first approach and maps the observation onto a latent feature space where the Kalman update is feasible. To overcome the nonlinearity, RKNs maintain a bank of base linear systems that can be interpolated over time. An open question for RKN is how to consider the several possible evolution trends of the future. In general, our contributions are as follows:
\begin{enumerate}
    \item We present an alternative approach for future trajectory prediction that accounts for multimodality and uncertainty. In particular, we employ the novel Recurrent Kalman Network \citep{becker2019recurrent} with variational inference technique to introduce a deep learning model that can model multimodal dynamics. Our model enjoys the interpretability of a state-space model while scaling well for real-time inference and prediction tasks.
    \item We demonstrate the proposed models on a real-world task, which is to model taxi trajectory data. Traffic forecasting is an inspiring problem in autonomous driving because of its nonlinear temporal and spatial dependency. Understanding this traffic behavior is important for monitoring urban traffic and electronic traffic dispatching.
\end{enumerate}
\section{Related Works}
In machine learning, the Bayesian framework is often employed to quantify the degree of uncertainty in an event. In Bayesian modeling, probabilities are adopted to systematically reason about model uncertainty \citep{murphy2012machine}.  A prominent example of combining Bayesian modeling and deep learning are variational autoencoders (VAEs) \citep{kingma2013auto, rezende2014stochastic}. They are unsupervised deep learning models which attempt to find a compressed representation of the observations in some latent space. The VAEs have enjoyed widespread adoption and have been extended to incorporate temporal dependencies. 

Time series data are often described by state-space models (SSMs). State-space models assume that there is an underlying system that governs the observation generation process. This system evolves over time, causing temporal dependencies in the observations. In state-space models, both the observations and the underlying system states are modeled with probability distributions. The notion of the state-space model has its origin back to the 1960s, with the introduction of the Kalman Filter for linear and Gaussian system \citep{kalman1960new}. Despite its elegant computation and simplicity, the Kalman Filter is limited to linear and Gaussian state-space models. A line of works in the control theory community proposes to address multimodality and nonlinearity problems by maintaining a bank of $K$ linear systems and interpolate between them \citep{ackerson1970state, murphy1998switching,ghahramani2000variational,lee2004multimodal,fox2008nonparametric,oh2005variational}. However, these methods often require the knowledge of system parameters and are not designed to work with high-dimensional data. 

In the last few years, there have been several efforts made to provide deep state-space models \citep{becker2019recurrent, karl2016deep,fraccaro2017disentangled, rangapuram2018deep}. They enjoy tractability, but they are often not expressive enough to capture multimodality. Non-linear deep SSMs \citep{zheng2017state, doerr2018probabilistic,gedon2020deep,krishnan2017structured, kingma2013auto, rangapuram2018deep} have emerged as an alternative, but they lose their tractability and have to resort to approximation techniques. Although all these deep state-space models are successful in modeling complex real-world time series data, they are not explicitly designed to capture multimodality.

Some previous works have proposed methods to account for multimodality. \citep{qiu2020variational} introduces a novel inference technique that accounts for multimodality. Other works employ the idea of switching regimes incorporated with deep learning, such as \citep{johnson2016composing, farnoosh2020deep,dai2016recurrent,liu2018structured}. These models assume the Markov assumption on the state evolution. The Markov assumption has been relaxed in several other works by letting the switching variable depend on previous system state or observations \citep{linderman2017bayesian,becker2019switching,dong2020collapsed}. Another approach to introducing multimodality is to model the system dynamics as a Gaussian mixture model \citep{alspach1972nonlinear, wills2017bayesian, yu2012particle, huber2011adaptive}.

Recurrent Kalman Network \citep{becker2019recurrent} is a probabilistic recurrent neural network architecture for sequential data that employs Kalman updates to learn a latent state representation of the system. It achieves competitive results on various state estimation tasks while providing reasonable uncertainty estimates and efficiency. In this work, we propose to combine Recurrent Kalman Network with switching Kalman Filter to account for multimodal dynamics of time series data.
\section{Methodology}
\begin{figure}[t]
    \centering
    \includegraphics[width=0.89\textwidth]{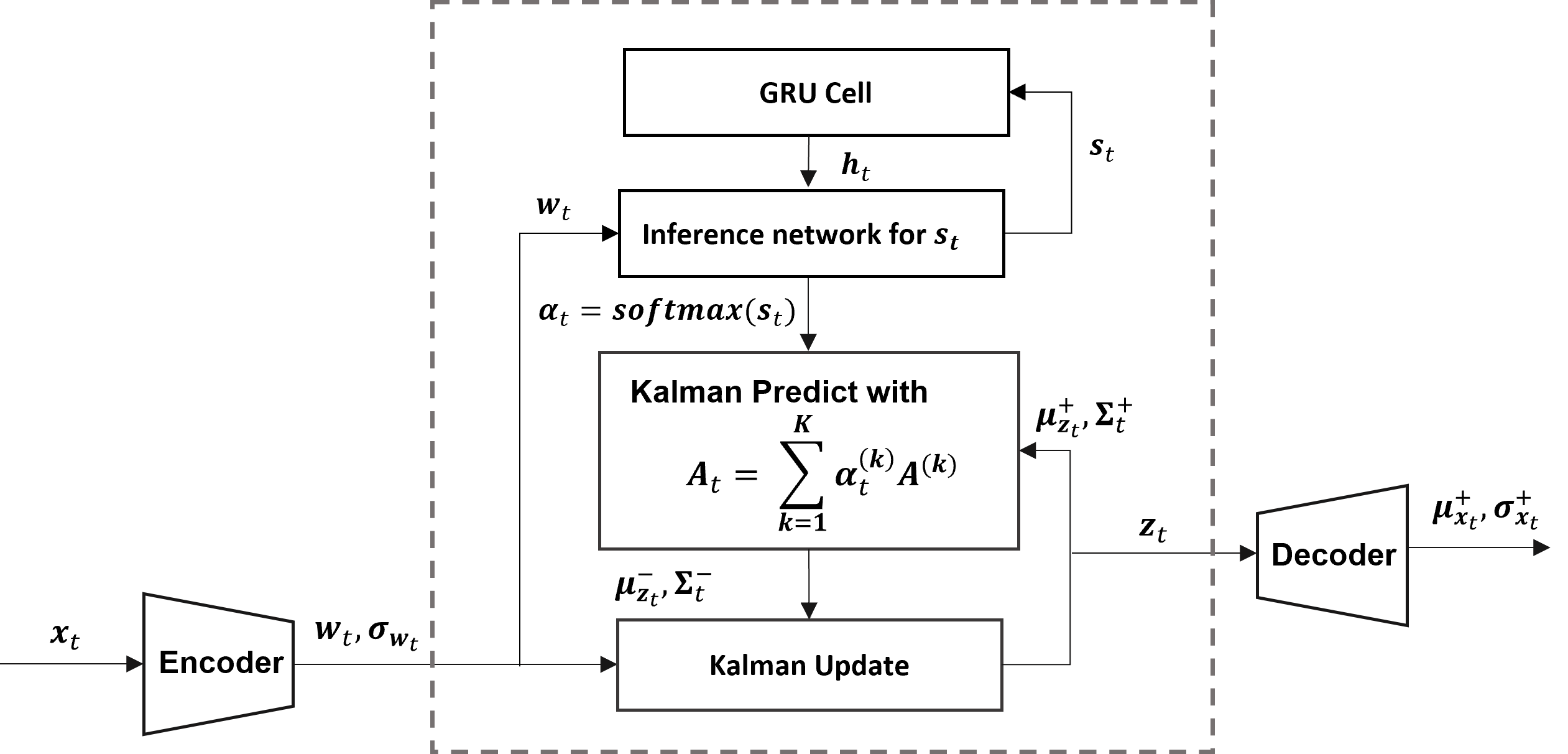}
    \caption{The architecture of the Switching Recurrent Kalman Network. The encoder maps the observations onto a latent feature space. The encoder also produces an uncertainty vector for the mapped latent observations. There is a gated recurrent unit cell that stores information about the switching variable over time. The latent observation is combined with the GRU cell to approximate the posterior distribution for the switching variable. A single sample of this posterior goes to a softmax layer to produce the weighting coefficients for the transition base matrices. The posterior distribution of the latent state from the previous time step is combined with the weighted base matrices to form the predictive distribution for the current latent state. The resulting prediction is then filtered using the latent observation and its uncertainty vector in the Kalman update step. After that, a single sample from the posterior is input to the decoder to parameterize the approximated distribution for the current observation.}
    \label{fig:srkn_arch}
\end{figure}

The Switching Recurrent Kalman Network (SRKN) is an extension of the Recurrent Kalman Network \citep{becker2019recurrent} that accounts for multimodality. The architecture of the model is visualized in Figure \ref{fig:srkn_arch}. The SRKN employs a latent observation and latent state space. The observations, such as images, are mapped onto a latent observation space where linear dynamics are feasible.  The transformation to this latent feature space is given by the SRKN encoder and can be learned end-to-end. In this latent space, exact posterior inference can be done with Kalman Filter. This idea was already adopted before \citep{fraccaro2017disentangled} to disentangle high-dimensional observations like images to a pseudo-observation latent space where linear assumption may apply.
\subsection{The Generative Model}
\begin{figure}[t]
    \centering
    \subfigure[The generative model]{\label{fig:example}\includegraphics[width=0.26\textwidth]{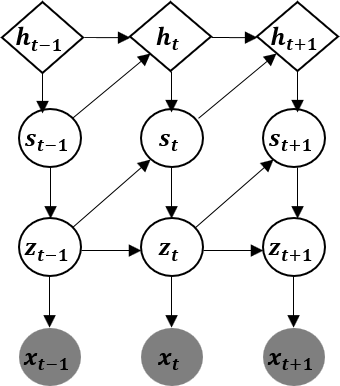}}    
        \hspace{1cm}      
     \subfigure[The inference model]{\label{fig:example}\includegraphics[width=0.26\textwidth]{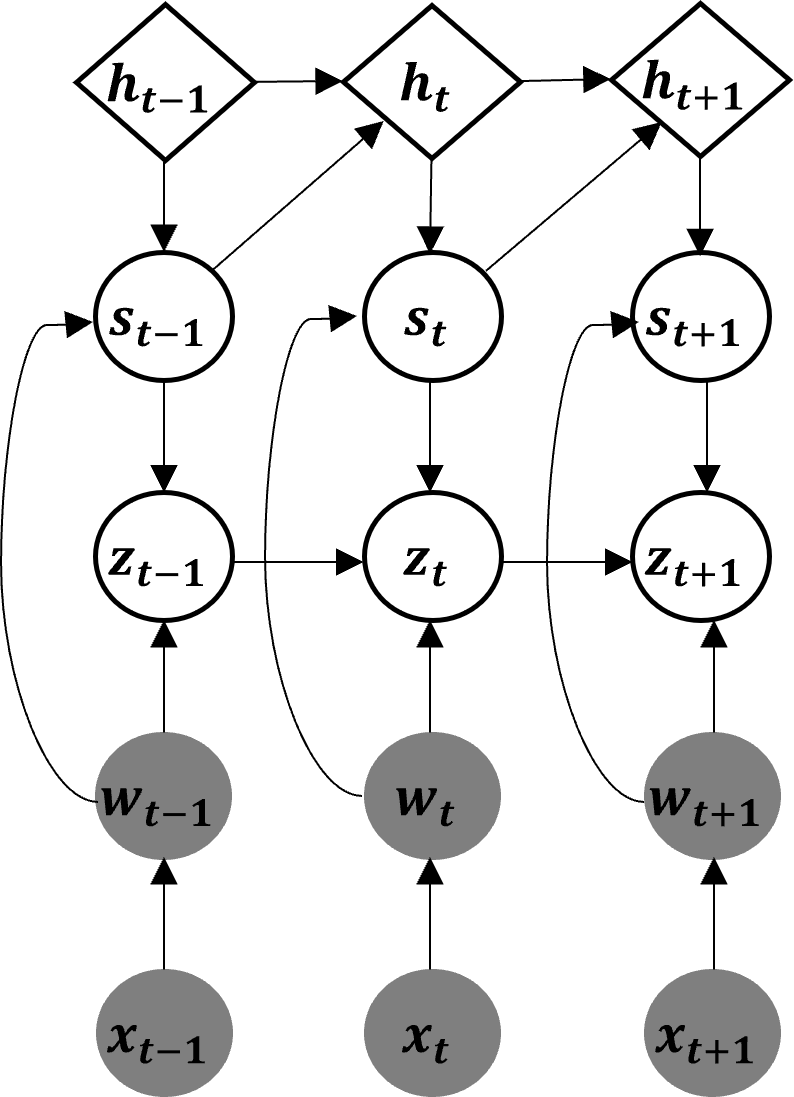}}    
    \caption{The graphical model of the Switching Recurrent Kalman Network. In the generative model, the switching variable $\mathbf{s}_t$ is conditioned on its distribution up to the current time step and the previous latent state $\mathbf{z}_t$. The deterministic recurrent cell $\mathbf{h}_t$ stores information about $\mathbf{s}_t$ over time. $\mathbf{s}_t$ determines the weights of the base matrices. The linear model in time step $t$ is a weighted sum of the base systems. The current latent state is related to the previous latent state by a linear model, given the switching variable. 
    The observation $\mathbf{x}_t$ is disentangled from the latent state. In the inference model, the dependency of $\mathbf{s}_t$ on $\mathbf{z}_{t-1}$ is discarded.  In addition, the real observations are mapped onto a latent representation $\mathbf{w}_t$. $\mathbf{w}_t$ is used to do the inference of $\mathbf{s}_t$ and $\mathbf{z}_t$. This has the advantage that the inference of $\mathbf{z}_t$ is available in closed-form with the Kalman Filter.
    }
    \label{fig:srkn}
\end{figure}
\paragraph{The Generative Model in the Latent Space.} The latent state space $\mathcal{Z} = \mathbb{R}^{2m}$ is related to the latent observation by a simple linear emission function:
\begin{equation}
 \mathbf{w}_t = \mathbf{H}\;\mathbf{z}_t ;  \: \: \: \: \; \; \mathbf{H} = [ \mathbf{I}_m \: \: \: \:  \mathbf{0}_{m\times m}],
\end{equation}
where $m$ is the dimensionality of the latent observation, $\mathbf{I}_m$ denotes the identity matrix, and $\mathbf{0}_{m\times m}$ represents a $m\times m$ matrix filled with zeros. This emission model effectively splits the latent state vector into two parts. The first (upper) part contains information which is included in the observation, and the second (lower) part, the memory, is the information inferred over time, e.g., velocities. Depending on the input dimension (images or real-valued), an uncertainty vector is also output by the decoder.
\paragraph{The Generative Model in the Observation Space.} The decoder $f_{dec}$ parameterizes the distribution of the reconstructed observation using a single sample of the latent state:

\begin{equation}
    p(\mathbf{x}_t|\mathbf{z}_t, \mathbf{s}_t) = \mathcal{N} (\boldsymbol{\mu}_{\mathbf{x}_t}, \boldsymbol{\Sigma}_{\mathbf{x}_t})\;\: \textrm{where} \;\:[ \boldsymbol{\mu}_{\mathbf{x}_t}, \boldsymbol{\Sigma}_{\mathbf{x}_t}] = f_{dec} (\mathbf{z}_t) \;\:;\;\: \mathbf{z}_t \sim p(\mathbf{z}_t|\mathbf{s}_t, \mathbf{z}_{t-1})
\end{equation}
\paragraph{The Transition Model.}The SRKN assumes the system dynamics evolve locally linearly over time. This way, the system state can be inferred online with the Kalman Filter \citep{kalman1960new}. To obtain a locally linear transition dynamics, the SRKN maintains a bank of transition base matrices $\mathbf{A}^{(k)}$, and the transition matrix at each time step is a weighted sum of these base matrices. The predictive distribution for the latent state at time step $t$ is
\begin{equation}
\begin{aligned}
    \mathbf{A}_t = \sum^K_{k=1} \alpha^{\left ( k \right )}_t\mathbf{A}^{\left ( k \right )} \:\:\: ; \:\:\:  \boldsymbol{\alpha_t} = (\alpha^{\left (1 \right )}_t, ..., \alpha^{\left ( K \right )}_t) = softmax(\mathbf{s}_t)\:\:\: ; \:\:\: \sum^K_{k=1} \alpha^{\left ( k \right )}_t = 1\:\:\: ; \:\:\:  \alpha_t^{(k)}\geq 0\\
    p(\mathbf{z}_t|\mathbf{s}_{t}, \mathbf{z}_{t-1}) = \mathcal{N} (\mathbf{\boldsymbol{\mu}}_{\mathbf{z}_t^-}, \boldsymbol{\Sigma}_{\mathbf{z}_t^-}) \: \: \: \textrm{where} \: \: \: \mathbf{\mu}_{\mathbf{z}_t^{-}} = \textbf{A}_t \boldsymbol{\mu}_{\mathbf{z}_{t-1}^{+} }\: \: ; \: \: 
\boldsymbol{\Sigma}_{\mathbf{z}_t^{-}} = \textbf{A}_t\boldsymbol{\Sigma}_{\mathbf{z}_{t-1}^{+}}\textbf{A}_t^T + \textbf{I}.\boldsymbol{\sigma}^{trans}.
\end{aligned}
\end{equation}
Here $\boldsymbol{\mu}_{\mathbf{z}_t^{-}}$ and $\boldsymbol{\Sigma}_{\mathbf{z}_t^-}$ denote the prior mean and the prior covariance of $\mathbf{z}_t$ while $\boldsymbol{\mu}_{\mathbf{z}_{t-1}^{+}}$ and  $\boldsymbol{\Sigma}_{\mathbf{z}_{t-1}^+}$ represents the mean and the covariance of the posterior of the previous latent state $\mathbf{z}_{t-1}$. In addition, $\boldsymbol{\alpha}_{t}^{(k)}$ indicates the weight assigned to the $k$-th linear base matrix. Its value is non-negative and all weights sum to one. The idea of having several transition base matrices is close to the Switching Kalman Filter \citep{murphy1998switching}. 

The weights assigned to the transition base matrices are given by the switching variable $\mathbf{s}_t$. This switching variable is conditioned on its distribution in previous time steps and on the latent state of the previous time step. To this extend, a gated recurrent unit $g$ is adopted to store information about the switching variable over time. A neural network $f_{trans}$ is used to combine information from the latent state and the switching variable
\begin{equation}
\begin{aligned}
    p(\mathbf{s}_t|\mathbf{s}_{<t}, \mathbf{z}_{t-1}) = \mathcal{N} (\boldsymbol{\mu}_{\mathbf{s}_t}, \mathbf{\Sigma}_{\mathbf{s}_t}) \: \: \: \textrm{where} \: \: \: [\boldsymbol{\mu}_{\mathbf{s}_t}, \mathbf{\Sigma}_{\mathbf{s}_t}] = f_{trans} (\mathbf{h}_{t}, \mathbf{z}_{t-1});\mathbf{h}_t = g(\mathbf{h}_{t-1}, \mathbf{s}_{t-1}) \\
   \boldsymbol{ \alpha}_t = softmax(\mathbf{s}_t) \: \: \: ;  \mathbf{s}_t \sim \mathcal{N} (\boldsymbol{\mu}_{\mathbf{s}_t}, \mathbf{\Sigma}_{\mathbf{s}_t}).
\end{aligned}
\end{equation}
The weighting coefficients for the base matrices are obtained by putting a sample of $ \mathbf{s}_t$ through the softmax layer.   
 In summary, the generative model is factorized as follows
 \begin{equation}
    p(\mathbf{x}_{1:T}, \mathbf{z}_{1:T}, \mathbf{s}_{1:T}) = \prod_{t=1}^T p(\mathbf{x}_t | \mathbf{s}_t, \mathbf{z}_t) p(\mathbf{z}_t | \mathbf{s}_t,  \mathbf{z}_{t-1}) p(\mathbf{s}_t | \mathbf{s}_{<t}, \mathbf{z}_{t-1}).
\end{equation}
\subsection{The Inference Model}
We propose the following factorization of the inference model:
\begin{equation}
\begin{aligned}
       q( \mathbf{s}_{1:T}, \mathbf{z}_{1:T} | \mathbf{x}_{1:T}) = \prod_{t=1}^T q(\mathbf{z}_t | \mathbf{s}_t, \mathbf{z}_{t-1}, \mathbf{x}_t)q(\mathbf{s}_t | \mathbf{s}_{<t}, \mathbf{x}_t)\\
 q(\mathbf{s}_t | \mathbf{s}_{<t}, \mathbf{x}_t) = \mathcal{N}(\boldsymbol{\mu}_{\mathbf{s}_t}, \boldsymbol{\Sigma}_{\mathbf{s}_t}) \: \: \: \textrm{where} \: \: \:  [\boldsymbol{\mu}_{\mathbf{s}_t}, \boldsymbol{\Sigma}_{\mathbf{s}_t}] = f_{inf}(\mathbf{s}_{<t}, \mathbf{x}_t)\\
   p(\mathbf{z}_t|\mathbf{s}_{t}, \mathbf{z}_{t-1}, \mathbf{x}_t)  = \mathcal{N} (\mathbf{\mu}_{\mathbf{z}_t^+}, \mathbf{\Sigma}_{\mathbf{z}_t^+})  \: \: \: \textrm{where} \: \: \:  [\mathbf{\mu}_{\mathbf{z}_t^+}, \mathbf{\Sigma}_{\mathbf{z}_t^+}] = Kalman\_Update (\boldsymbol{\mu}_{\mathbf{z}_t^-}, \boldsymbol{\Sigma}_{\mathbf{z}_t^-}).
       \end{aligned}
\end{equation}
The inference for $\mathbf{z}_t$ is given by the factorized Kalman update introduced by the RKN. Details about the factorized inference can be found in \citep{becker2019recurrent}. Here, the condition of $\mathbf{s}_t$ on $\mathbf{z}_{t-1}$ is discarded, see Figure \ref{fig:srkn}. Our empirical experiments show that removing this condition in the inference model resolves the mode averaging problem when training the model. Many previous approaches also omit some of the conditions in their inference models, see \citep{bayer2014learning, chung2015recurrent, li2018disentangled}.

The inference of the switching variable is done with the amortized variational inference technique \citep{gershman2014amortized}, where the inference networks and the generative networks are trained together. These networks have the task of parametrizing the probability distributions of the switching variable and the observations. Besides, the inference of the latent system state follows the elegant computational structure of the RKN, where the filtering process can be simplified to scalar operations.

\subsection{The Evidence Lower Bound}
Our model belongs to the class of variational approach \citep{jordan1999introduction}. The variational inference technique formulates a tractable lower bound for the complex distribution of interest and thus transforms the approximation of some intractable posterior into an optimization problem. This is obtained by finding an approximated posterior distribution that minimizes the KL-divergence \citep{kullback1997information, kullback1951information} of it to the real posterior. Minimizing the KL divergence is equivalent to maximizing the following evidence lower bound (ELBO):
\begin{equation}
    \begin{aligned}
    \mathcal{L}_{ELBO} &= \sum^T_{t=1}  \mathbb{E}_{q(\mathbf{z}_t | \mathbf{s}_t, \mathbf{z}_{t-1}, f_{\mathbf{w}}(\mathbf{x}_t)) q(\mathbf{s}_t | \mathbf{s}_{<t}, f_{\mathbf{w}}(\mathbf{x}_t))}[\log p(\mathbf{x}_t | \mathbf{s}_t, \mathbf{z}_t)]\\
    &- \mathbb{E}_{ q(\mathbf{s}_{t} | \mathbf{s}_{<{t}}, \mathbf{z}_{{t}-1}, f_{\mathbf{w}}(\mathbf{x}_{t-1}))}[\mathbb{E}_{q(\mathbf{z}_{{t-1}} | \mathbf{s}_{{t-1}}, \mathbf{z}_{{{t-2}}}, f_{\mathbf{w}}(\mathbf{x}_{{t-1}}))} \\& [\textrm{KL}(q(\mathbf{z}_t | \mathbf{s}_t, \mathbf{z}_{t-1}, f_{\mathbf{w}}(\mathbf{x}_t)|| p(\mathbf{z}_t|\mathbf{s}_t,\mathbf{z}_{t-1}))]]\\
    &- \mathbb{E}_{q(\mathbf{s}_1| f_{\mathbf{w}}(\mathbf{x}_1))}[...\mathbb{E}_{q(\mathbf{s}_{{t}} | \mathbf{s}_{<{{t}}}, \mathbf{z}_{{{t}}-1}, f_{\mathbf{w}}(\mathbf{x}_{{t}}))}[\mathbb{E}_{q(\mathbf{z}_{{t-1}} | \mathbf{s}_{{t-1}}, \mathbf{z}_{{{t}}-2}, f_{\mathbf{w}}(\mathbf{x}_{{t-1}}))}\\ & [\textrm{KL}(q(\mathbf{s}_t | \mathbf{s}_{<t}, \mathbf{z}_{t-1}, f_{\mathbf{w}}(\mathbf{x}_t))|| p(\mathbf{s}_t | \mathbf{s}_{<t}, \mathbf{z}_{t-1}))]]].
    \end{aligned}
\end{equation}

Here, $f_{\mathbf{w}}$ denotes the function that maps the real observation $\mathbf{x}_t$ to the latent observation $\mathbf{w}_t$. The derivation for this ELBO is given in Appendix \ref{appendix:elbo}. We introduce a scaling factor for each component of the ELBO. These scaling factors are motivated by the $\beta$-VAE \citep{higgins2016beta} and govern the trade-off between the reconstruction term and the regularization term. Depending on the problems at hand, tuning these scaling factors might be beneficial to the overall training performance. Besides, we add a prediction loss term to guide the model training process. This prediction loss term is the weighted sum of $K$ observation probabilities. Each probability $p^{(k)}(\mathbf{x}_t|\mathbf{s}_{t},\mathbf{z}_{t-1})$ refers to the observation probability when the transition of the latent state $\mathbf{z}_t$ follows the linear base system $\boldsymbol{A}^{(k)}$.  Intuitively, the prediction loss term corresponds to the log probability of a mixture model with $K$ components. The prediction loss term enforces the model to assign higher weight on the base systems that are more likely to generate the subsequent observation. The model is learned end-to-end from data by maximizing the following objective function:
\begin{equation}
    \mathcal{L}_{Objective} =  \mathcal{L}_{\beta\_ ELBO} +      \beta_{pred}\mathcal{L}_{Pred},
\end{equation}
where
\begin{align}
    \mathcal{L}_{pred} &= \sum_{t=1}^T \log \sum_{k=1}^K \alpha_{t}^{(k)}p^{(k)}(\mathbf{x}_t|\mathbf{s}_{t},\mathbf{z}_{t-1})\nonumber\\&
    \textrm{where}\; p^{(k)}(\mathbf{x}_t|\mathbf{s}_{t},\mathbf{z}_{t-1}) = \mathbb{E}_{p^{(k)}(\mathbf{z}_t|\mathbf{s}_{t},\mathbf{z}_{t-1})}[p(\mathbf{x}_t|\mathbf{s}_{t},\mathbf{z}_{t}) p^{(k)}(\mathbf{z}_t|\mathbf{s}_{t},\mathbf{z}_{t-1})] \\& p^{(k)}(\mathbf{z}_t|\mathbf{s}_{t},\mathbf{z}_{t-1}) = \mathcal{N}(\mathbf{z}_t; \mathbf{A}^{(k)}\mathbf{z}_{t-1}, \mathbf{A}^{(k)}\boldsymbol{\Sigma}_{\mathbf{z}_{t-1}}(\mathbf{A}^{(k)} )^T).
\end{align}
$\mathcal{L}_{\beta\_ ELBO}$ refers to the ELBO where the reconstruction loss term, the KL-divergence for $\mathbf{z}_t$ and the KL-divergence for $\mathbf{s}_t$ have a scaling factor $\beta_{rec}$, $\beta_{\mathbf{z}}$ and $\beta_{\mathbf{s}}$, respectively. 
\section{Experiments}
In this section, we evaluate the SRKN with several data sets. We first consider a simulated 2-d time series data set whose dynamics have four modes and a synthetic image data set of car motions that follow an underlying structure. We further apply the SRKN to the real-world taxi data set \citep{ecmlpkdd2015online}. The results are compared against several methods for modelling time-series data, including the RKN \citep{becker2019recurrent}, VRNN-GMM \citep{chung2015recurrent}, VDM \citep{qiu2020variational}, DMM-IAF \citep{krishnan2015deep, kingma2016improved}.

\subsection{Evaluation metrics}
We choose four metrics to evaluate the predictions quantitatively. They include i) one-step prediction loss $\log p(\mathbf{x}_t|\mathbf{x}_{<t})$, ii) multi-step prediction loss $\log p(\mathbf{x}_{t:t+\tau}|\mathbf{x}_{<t})$, iii) reconstruction log likelihood $\log p(\mathbf{x}_t|\mathbf{x}_{\leq t})$ and iv) Wasserstein distance \citep{villani2009optimal}. 
A real-valued observation is modeled with a multivariate Gaussian distribution with diagonal covariance. The negative Gaussian reconstruction log-likelihood for a sequence in this case is
\begin{equation}
    \mathcal{L}(\boldsymbol{x}_{1:T})  = \frac{1}{T} \sum^T_{t=1} - \log \mathcal{N}(\boldsymbol{x}_t|\boldsymbol{\mu}_{\boldsymbol{x}_t}^+, \boldsymbol{\sigma}^+_{\boldsymbol{x}_t}).
\end{equation}
The negative high-dimensional data are modeled with a Bernoulli distribution. The reconstruction log-likelihood is computed as follows
\begin{equation}
        \mathcal{L}(\boldsymbol{x}_{1:T})  = - \frac{1}{T} \sum^T_{t=1} \sum_{d=0}^D \boldsymbol{x}^{(d)}_t \log(\boldsymbol{\mu}_{\boldsymbol{x}_t}^{(d)+}) + (1-\boldsymbol{\mu}_{\boldsymbol{x}_t}^{(d)+}) \log(1-\boldsymbol{\mu}_{\boldsymbol{x}_t}^{(d)+}).
\end{equation}
The one-step prediction loss term demonstrates the prediction power of the model for the next time step, given the observations up to the current time step
\begin{equation}
   \mathcal{L}_{one\_ step}(\boldsymbol{x}_{1:T}) = \sum^{T-1}_{t=1}-\log p(\mathbf{x}_{t+1}|\mathbf{x}_{1:t}).
\end{equation}
To compute the multi-step prediction loss, we generate $n=100$ predictions for the rest of the sequence, given observations up to time step $\tau$
\begin{equation}
     \mathcal{L}_{multi\_steps}(\boldsymbol{x}_{1:T}) = \frac{1}{n}\sum^n_{i=1} \sum^{T-1}_{t=\tau}- \log p_{(i)}(\mathbf{x}_{t+1}|\mathbf{x}_{1:\tau}).
\end{equation}
The Wasserstein distance accounts for both diversity and accuracy of prediction. To approximate the Wasserstein distance, we select $n = 100$ samples from the test set that have similar initial trajectories. The model is expected to generate sample predictions that match all ground truth continuations in the test set, given the initial trajectories. We refer to \citep{villani2009optimal, qiu2020variational} for the details of the Wasserstein distance.
\subsection{Toy Experiments}
\begin{figure}[t]
    \centering
    \begin{minipage}[t]{0.45\textwidth}
    \centering
    \vspace{2mm}
    \subfigure[]{\label{fig:2mode2dim_gen}\includegraphics[width=\textwidth]{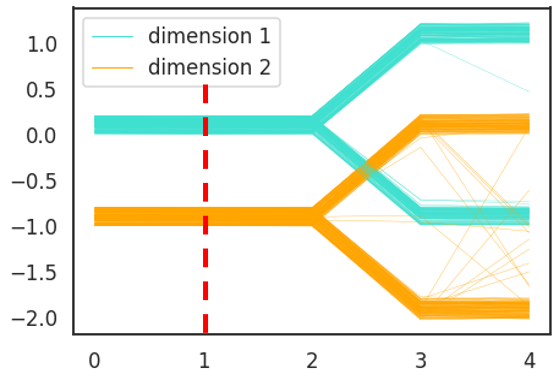}}
    \end{minipage}%
    \begin{minipage}[t]{0.5\textwidth}
    \centering
      \subfigure[]{\label{fig:2mode2dim_ex1}\includegraphics[width=0.5\textwidth]{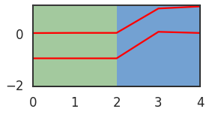}}%
     \subfigure[]{\label{fig:2mode2dim_ex2}\includegraphics[width=0.5\textwidth]{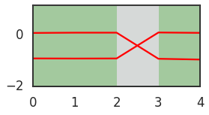}}
     \subfigure[]{\label{fig:2mode2dim_ex3}\includegraphics[width=0.5\textwidth]{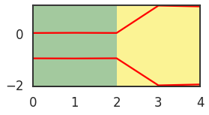}}%
     \subfigure[]{\label{fig:2mode2dim_ex4}\includegraphics[width=0.5\textwidth]{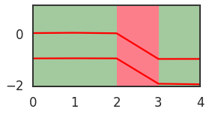}} 
    \end{minipage}
    \caption{Figure (a): Generated trajectories by the SRKN. Figures (b-e): The model assigns different transition modes to each possible continuation of the trajectory. Each color corresponds to one transition dynamic mode. Each time step is color-coded with the mode that the model assigns the highest weight to. The first two time steps are given (indicated by the dashed red line), and the model was asked to predict the next 3 time steps.  }
    \label{fig:2mode2dim_gen_hmsrkn}
\end{figure}

\begin{figure}[t]
    \centering
    \includegraphics[width=0.98\textwidth]{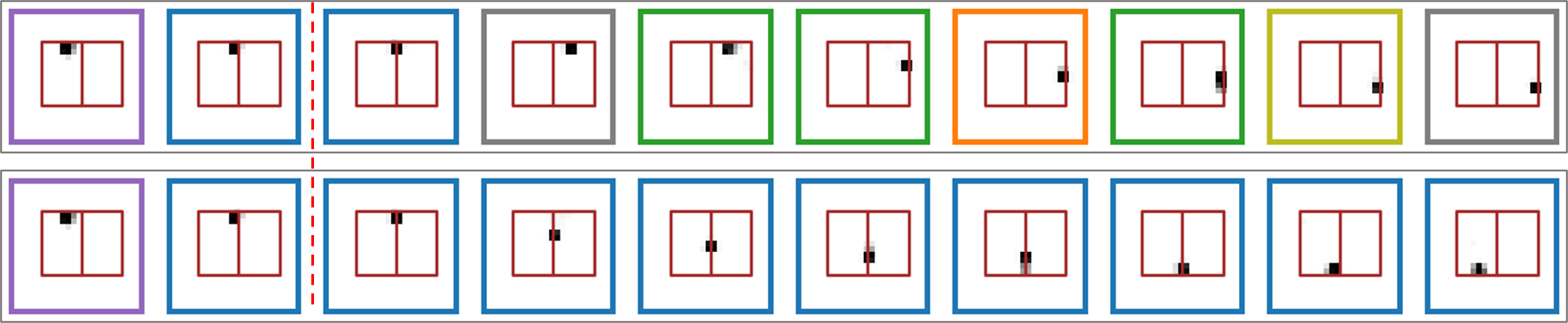}
    \caption{Two image sequences were generated by the SRKN given the two first time steps. Each color corresponds to one transition dynamic mode. Each image is color-coded with the mode that the model assigns the highest weight to. The two rectangles are not present in the dataset but only serve for visualization. Here, the model can determine the two potential trajectories that the car can follow when approaching the crossroad. }
    \label{fig:square_high_example_color}
\end{figure}

\paragraph{2-d Synthetic Data Set.} We start with a simple two-dimensional data set to verify the ability of the proposed model in capturing multimodality. Each sequence consists of five time-steps. The data sequences have a constant value in the first three steps. At time step $4$, each dimension of the data point can switch to two possible modes, causing the data to have four modes in total. We visualize the results in Figure \ref{fig:2mode2dim_gen_hmsrkn}. The model can successfully capture the switching point at the fourth time step.
\paragraph{Synthetic Car Trajectories Images Data Set.}  Next, we evaluate the SRKN on a simple synthetic car trajectories image dataset. The observations here are sequences of images of $24 \times 24$ pixels. The black square represents a car whose trajectory follows an underlying pattern containing two rectangles next to each other. Each image illustrates the position of the car at a time. The car never goes in the opposite direction at any given time step. The qualitative results are demonstrated in Figure \ref{fig:square_high_example_color}. Each image is color-coded with the dominant mode that the model predicts. The black square seems blurred in the later time steps, which is presumably caused by the transition noise incorporated in the model. It is noteworthy that although the model was trained on sequences of only length $6$, they can give good predictions for longer sequences. In other words, the model can learn and generalize the underlying dynamics of the data. Hence, a potential application of the SRKN is to model real-world trajectories image data in autonomous driving. Note that the two rectangles are not included in the dataset but only serve evaluation purposes. 

The quantitative results for the toy experiments are given in Table \ref{table:toy}. Our model achieves competitive results as the VDM on the four mode data set, while on the pendulum image data set, it achieves the best one-step prediction, multi-step prediction and Wasserstein distance.

\begin{table}[t] 
\label{table:toy}
\centering
\begin{tabular}{lcccc|cccc}
\toprule
\multicolumn{1}{c}{} & \multicolumn{4}{c}{Four modes data set}                         & \multicolumn{4}{c}{Car trajectories data set}                \\ \midrule
                       & 1-step       & Multi-step    & w-dist    & LL             & 1-step      & Multi-step    & w-dist    & LL            \\ \midrule
VDM     &  \textbf{-4.83}          & \textbf{2.11}      & \textbf{0.10}          & -4.90          & 7.04          & 7.45          & 6.44          & 6.23          \\           
RKN       & -3.91          & 3.41          & 0.22          & -4.83          & \textbf{4.33} & 5.33          & 7.11          & \textbf{2.63} \\
VRNN                & -3.96 & 2.59 & 0.13 & -5.06          & 5.14          & 8.14          & 6.21          & 4.93          \\
DMM         & -2.94          & 4.00          & 0.72          & \textbf{-5.21} & 7.86          & 8.04          & 6.44          & 6.87          \\           
SRKN                   & -4.12          & 2.37          & \textbf{0.10} & -5.07          & \textbf{4.33} & \textbf{5.10} & \textbf{4.40} & 2.74          \\ 
\bottomrule
\end{tabular}
 \vspace{0.5cm}  
\caption{Quantitative results on four modes and car trajectories datasets. In the four modes data set, the SRKN and the VDM have the smallest Wasserstein distance. This indicates their similar performance in prediction and capturing multimodality. Compared to the RKN, the SRKN achieves a smaller one-step and multi-step prediction loss. Among all baselines, only the VDM has a better one-step and multi-step prediction loss than the SRKN. In the car trajectories data set, the SRKN outperforms all the baseline models in terms of prediction loss and Wasserstein distance. The reconstruction loss of the RKN in this image dataset is slightly better than the SRKN.}
\end{table}

\subsection{Real World Taxi Data Set.} 
\begin{figure}[t]
    \centering
    \includegraphics[width=0.97\textwidth]{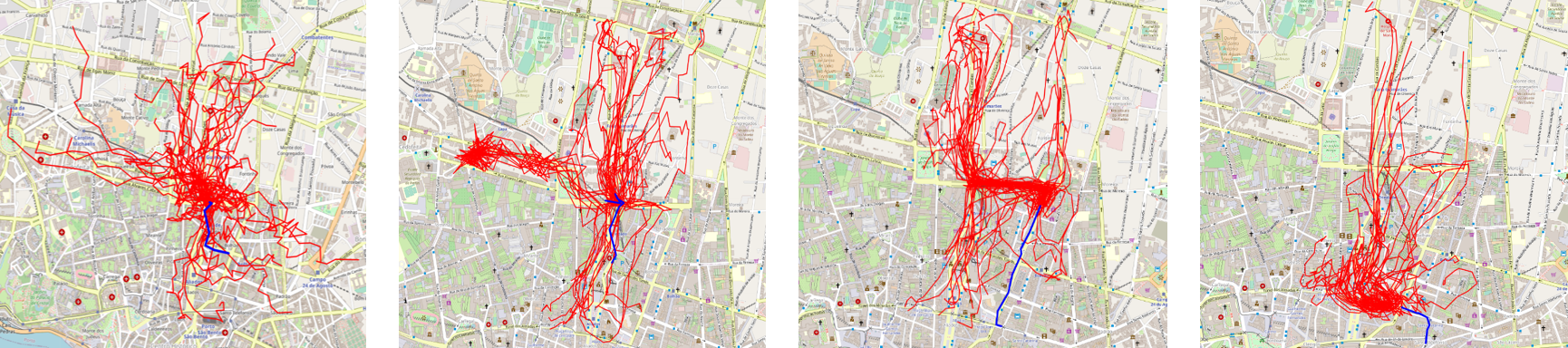}
    \caption{50 generated trajectories (red) given the initial observations (blue). The model can generate trajectories that follow the general evolving structure of the underlying map. However, the model cannot capture the underlying map well.}
    \label{fig:taxi_srkn}
\end{figure}

To validate the effectiveness of the proposed model, we experiment on the Porto taxi data set. The original data set consists of 1.7 million records, coming from 442 taxis running in Porto, Portugal. For evaluation, we reuse the preprocessing pipeline suggested in \citep{qiu2020variational}. We select only the trajectories in the city area and only extract the first $30$ time steps. The resulting dataset is split into the training set of size $86 386$, the validation set of size 200, and the test set of size $10 000$. Figure \ref{fig:taxi_srkn} demonstrates the qualitative forecasting results. The task is to predict the next $20$ time steps given the first $10$ time steps. The model can capture the multimodal dynamics and give predictions that follow the underlying evolution structure of the map. Compared to the state-of-the-art model for multimodality such as the VDM, the SRKN cannot achieve such good prediction results. This could be because while SRKN employs a linear state transition model, the state transition in the VDM is nonlinear and is represented by a powerful deep neural network. 
\begin{table}[!ht]
 \centering
\begin{tabular}{lccccc}
\toprule
\multicolumn{1}{c}{} & \multicolumn{5}{c}{Taxi data set}                                                                   \\ \midrule
                       & 1-step       & Multi-step    & w-dist    & LL           & \multicolumn{1}{l}{\# parameters}  \\ \midrule
VDM                    & \textbf{-3.68} & \textbf{2.88} & \textbf{0.59} & -4.33          & \multicolumn{1}{c}{\textbf{22056}} \\
RKN                    & -2.9           & 4.2           & 2.07          & -4.25          & \multicolumn{1}{c}{23118}          \\
VRNN               & -2.77          & 5.51          & 2.43          & -4.09          & \multicolumn{1}{c}{22352}          \\
DMM                & -2.45          & 3.29          & 0.70           & \textbf{-4.35} & \multicolumn{1}{c}{22248}          \\
SRKN                   & -2.35          & 3.16          & 0.75          & -4.34          & \multicolumn{1}{c}{33742}          \\ \bottomrule

\end{tabular}
 \vspace{0.5cm}  
\caption{Quantitative results on taxi data set. The VDM outperforms all baseline models in terms of prediction loss and Wasserstein distance. In comparison to the RKN, the SRKN exhibits a much smaller Wasserstein distance and multi-step prediction loss. This shows an improvement of the SRKN compared to the RKN in the long-term and multimodal predictive power.}
\end{table}
\section{Conclusion}
We proposed a switching recurrent Kalman network for multimodal modeling of time series data. The model consists of a recurrent neural network for the switching variable and a locally linear state transition model. It operates on a latent observation space where a linear transition model is feasible. This enforces the state-space model assumption and enjoys an explicit notion of the system state. The inference of the system state follows the efficient computation structure of the RKN, while the inference of the switching variable is performed using amortized variational inference method. The model shows the ability to capture multimodality on the real-world Porto taxi trajectories dataset. Besides, our model enjoys the interpretability of a state-space model with switching regimes and outperforms the baseline models on high-dimensional car trajectory data. The ability of our model to incorporate uncertainty and multimodality in future predictions promises a wide range of applications in autonomous driving, such as the trajectory prediction of pedestrians and nearby vehicles.  
\bibliographystyle{abbrvnat}%
\bibliography{bibliography}


\appendix

\section{Appendix}
\input{elbo}

\end{document}

%% file: elbo.tex
\subsection{ELBO Derivation for the Switching Recurrent Kalman Filter} \label{appendix:elbo}
In this section, a lower bound to the marginal likelihood $p(\mathbf{x}_{1:T})$ is derived
\begin{equation}
\begin{aligned}
\textrm{KL} [q(\mathbf{s}_{1:T}, \mathbf{z}_{1:T} | \mathbf{x}_{1:T}) || p(\mathbf{s}_{1:T},\mathbf{z}_{1:T} | \mathbf{x}_{1:T})]
\\= \mathbb{E}_{q(\mathbf{s}_{1:T}, \mathbf{z}_{1:T} | \mathbf{x}_{1:T})}[\log q(\mathbf{s}_{1:T}, \mathbf{z}_{1:T} | \mathbf{x}_{1:T}) - \log p(\mathbf{s}_{1:T},\mathbf{z}_{1:T} | \mathbf{x}_{1:T})]
\\= \mathbb{E}_{q(\mathbf{s}_{1:T}, \mathbf{z}_{1:T} | \mathbf{x}_{1:T})}[\log q(\mathbf{s}_{1:T}, \mathbf{z}_{1:T} | \mathbf{x}_{1:T}) - \log p(\mathbf{s}_{1:T},\mathbf{z}_{1:T}, \mathbf{x}_{1:T}) - \log p(\mathbf{x}_{1:T})]
\\= \mathbb{E}_{q(\mathbf{s}_{1:T}, \mathbf{z}_{1:T} | \mathbf{x}_{1:T})}[\log q(\mathbf{s}_{1:T}, \mathbf{z}_{1:T} | \mathbf{x}_{1:T}) - \log p(\mathbf{s}_{1:T},\mathbf{z}_{1:T}, \mathbf{x}_{1:T})] - \mathbb{E}_{q(\mathbf{s}_{1:T}, \mathbf{z}_{1:T} | \mathbf{x}_{1:T})}[\log p(\mathbf{x}_{1:T})]
\\= \mathbb{E}_{q(\mathbf{s}_{1:T}, \mathbf{z}_{1:T} | \mathbf{x}_{1:T})}[\log q(\mathbf{s}_{1:T}, \mathbf{z}_{1:T} | \mathbf{x}_{1:T}) - \log p(\mathbf{s}_{1:T},\mathbf{z}_{1:T}, \mathbf{x}_{1:T})] - \log p(\mathbf{x}_{1:T})
\\= \mathcal{L}_{ELBO} -\log p(\mathbf{x}_{1:T})
\end{aligned}
\end{equation}
Since the KL divergence is a non-negative quantity, the term $\mathcal{L}_{ELBO}$ is a lower bound for the log likelihood of the observations $p(\mathbf{x}_{1:T})$. Next, we plug the generative and inference model in the ELBO:
\begin{equation}
\begin{aligned}
\mathcal{L}_{ELBO} &= \mathbb{E}_{q(\mathbf{s}_{1:T}, \mathbf{z}_{1:T} | \mathbf{x}_{1:T})}[\log q(\mathbf{s}_{1:T}, \mathbf{z}_{1:T} | \mathbf{x}_{1:T}) - \log p(\mathbf{s}_{1:T},\mathbf{z}_{1:T}, \mathbf{x}_{1:T})]\\
&=  \mathbb{E}_{q(\mathbf{s}_{1:T}, \mathbf{z}_{1:T} | \mathbf{x}_{1:T})}[\log \prod_{t=1}^T q(\mathbf{z}_t | \mathbf{s}_t, \mathbf{z}_{t-1}, f_{\mathbf{w}}(\mathbf{x}_t))q(\mathbf{s}_t | \mathbf{s}_{<t}, \mathbf{z}_{t-1}, f_{\mathbf{w}}(\mathbf{x}_t))\\
&- \log  \prod_{t=1}^T p(\mathbf{x}_t | \mathbf{s}_t, \mathbf{z}_t) p(\mathbf{z}_t | \mathbf{s}_t,  \mathbf{z}_{t-1}) p(\mathbf{s}_t | \mathbf{s}_{<t}, \mathbf{z}_{t-1})]\\
&= \mathbb{E}_{q(\mathbf{s}_{1:T}, \mathbf{z}_{1:T} | \mathbf{x}_{1:T})}[\sum_{t=1}^T \left (
\log q(\mathbf{z}_t | \mathbf{s}_t, \mathbf{z}_{t-1}, f_{\mathbf{w}}(\mathbf{x}_t)) + \log q(\mathbf{s}_t | \mathbf{s}_{<t}, f_{\mathbf{w}}(\mathbf{x}_t)) \right )  \\
&-  \sum_{t=1}^T (\log p(\mathbf{x}_t | \mathbf{s}_t, \mathbf{z}_t) +\log p(\mathbf{z}_t | \mathbf{s}_t,  \mathbf{z}_{t-1}) + \log p(\mathbf{s}_t | \mathbf{s}_{<t}, \mathbf{z}_{t-1}))]\\
&=\sum_{t=1}^T\mathbb{E}_{q(\mathbf{s}_{1:T}, \mathbf{z}_{1:T} | \mathbf{x}_{1:T})}[ \log p(\mathbf{x}_t | \mathbf{s}_t, \mathbf{z}_t)] \\
&+ \sum_{t=1}^T\mathbb{E}_{q(\mathbf{s}_{1:T}, \mathbf{z}_{1:T} | \mathbf{x}_{1:T})}[\log q(\mathbf{z}_t | \mathbf{s}_t, \mathbf{z}_{t-1}, f_{\mathbf{w}}(\mathbf{x}_t)) - \log p(\mathbf{z}_t|\mathbf{s}_t,\mathbf{z}_{t-1})] \\
&+\sum_{t=1}^T\mathbb{E}_{q(\mathbf{s}_{1:T}, \mathbf{z}_{1:T} | \mathbf{x}_{1:T})}[ \log q(\mathbf{s}_t | \mathbf{s}_{<t},  f_{\mathbf{w}}(\mathbf{x}_t))- \log p(\mathbf{s}_t | \mathbf{s}_{<t}, \mathbf{z}_{t-1})]
\end{aligned}
\end{equation}

Derivation of the evidence $\mathbb{E}_{q(\mathbf{s}_{1:T}, \mathbf{z}_{1:T} | \mathbf{x}_{1:T})}[ \log p(\mathbf{x}_t | \mathbf{s}_t, \mathbf{z}_t)]$
\begin{equation}
    \begin{aligned}
&    \mathbb{E}_{q(\mathbf{s}_{1:T}, \mathbf{z}_{1:T} | \mathbf{x}_{1:T})}[ \log p(\mathbf{x}_t | \mathbf{s}_t, \mathbf{z}_t)]\\ 
&= \int  \prod_{\tilde{t}=1}^T q(\mathbf{z}_{\tilde{t}} | \mathbf{s}_{\tilde{t}}, \mathbf{z}_{{\tilde{t}}-1}, f_{\mathbf{w}}(\mathbf{x}_{\tilde{t}}))q(\mathbf{s}_{\tilde{t}} | \mathbf{s}_{<{\tilde{t}}}, \mathbf{z}_{{\tilde{t}}-1}, f_{\mathbf{w}}(\mathbf{x}_{\tilde{t}})) \log p(\mathbf{x}_t | \mathbf{s}_t, \mathbf{z}_t)\\
&= \int q(\mathbf{z}_t | \mathbf{s}_t, \mathbf{z}_{t-1}, f_{\mathbf{w}}(\mathbf{x}_t)) q(\mathbf{s}_t | \mathbf{s}_{<t}, f_{\mathbf{w}}(\mathbf{x}_t)) \log p(\mathbf{x}_t | \mathbf{s}_t, \mathbf{z}_t)
\\
&= \mathbb{E}_{q(\mathbf{z}_t | \mathbf{s}_t, \mathbf{z}_{t-1}, f_{\mathbf{w}}(\mathbf{x}_t)) q(\mathbf{s}_t | \mathbf{s}_{<t}, f_{\mathbf{w}}(\mathbf{x}_t))}[\log p(\mathbf{x}_t | \mathbf{s}_t, \mathbf{z}_t)]
    \end{aligned}
\end{equation}
Derivation of the term $\mathbb{E}_{q(\mathbf{s}_{1:T}, \mathbf{z}_{1:T} | \mathbf{x}_{1:T})}[\log q(\mathbf{z}_t | \mathbf{s}_t, \mathbf{z}_{t-1}, f_{\mathbf{w}}(\mathbf{x}_t)) - \log p(\mathbf{z}_t|\mathbf{s}_t,\mathbf{z}_{t-1})]$:
\begin{equation}
    \begin{aligned}
   &\mathbb{E}_{q(\mathbf{s}_{1:T}, \mathbf{z}_{1:T} | \mathbf{x}_{1:T})}[\log q(\mathbf{z}_t | \mathbf{s}_t, \mathbf{z}_{t-1}, f_{\mathbf{w}}(\mathbf{x}_t)) - \log p(\mathbf{z}_t|\mathbf{s}_t,\mathbf{z}_{t-1})]\\
&= \int  \prod_{\tilde{t}=1}^T q(\mathbf{z}_{\tilde{t}} | \mathbf{s}_{\tilde{t}}, \mathbf{z}_{{\tilde{t}}-1}, f_{\mathbf{w}}(\mathbf{x}_{\tilde{t}}))q(\mathbf{s}_{\tilde{t}} | \mathbf{s}_{<{\tilde{t}}}, \mathbf{z}_{{\tilde{t}}-1}, f_{\mathbf{w}}(\mathbf{x}_{\tilde{t}})) \\& [\log q(\mathbf{z}_t | \mathbf{s}_t, \mathbf{z}_{t-1}, f_{\mathbf{w}}(\mathbf{x}_t)) - \log p(\mathbf{z}_t|\mathbf{s}_t,\mathbf{z}_{t-1})]\\&
= \int q(\mathbf{s}_{t} | \mathbf{s}_{<{t}}, \mathbf{z}_{{t}-1}, f_{\mathbf{w}}(\mathbf{x}_{t})) q(\mathbf{z}_{{t}} | \mathbf{s}_{{t}}, \mathbf{z}_{{{t}}-1}, f_{\mathbf{w}}(\mathbf{x}_{{t}}))q(\mathbf{z}_{{t-1}} | \mathbf{s}_{{t-1}}, \mathbf{z}_{{{t-2}}}, f_{\mathbf{w}}(\mathbf{x}_{{t-1}}))
\\& [\log q(\mathbf{z}_t | \mathbf{s}_t, \mathbf{z}_{t-1}, f_{\mathbf{w}}(\mathbf{x}_t)) - \log p(\mathbf{z}_t|\mathbf{s}_t,\mathbf{z}_{t-1})] \\&
= \mathbb{E}_{ q(\mathbf{s}_{t} | \mathbf{s}_{<{t}}, \mathbf{z}_{{t}-1}, f_{\mathbf{w}}(\mathbf{x}_{t-1}))}[\mathbb{E}_{q(\mathbf{z}_{{t-1}} | \mathbf{s}_{{t-1}}, \mathbf{z}_{{{t-2}}}, f_{\mathbf{w}}(\mathbf{x}_{{t}}))}[\textrm{KL}(q(\mathbf{z}_t | \mathbf{s}_t, \mathbf{z}_{t-1}, f_{\mathbf{w}}(\mathbf{x}_))|| p(\mathbf{z}_t|\mathbf{s}_t,\mathbf{z}_{t-1}))]]
    \end{aligned}
\end{equation}

Derivation of the term $\mathbb{E}_{q(\mathbf{s}_{1:T}, \mathbf{z}_{1:T} | \mathbf{x}_{1:T})}[ \log q(\mathbf{s}_t | \mathbf{s}_{<t},  f_{\mathbf{w}}(\mathbf{x}_t))- \log p(\mathbf{s}_t | \mathbf{s}_{<t}, \mathbf{z}_{t-1})]$
\begin{equation}
    \begin{aligned}
      & \mathbb{E}_{q(\mathbf{s}_{1:T}, \mathbf{z}_{1:T} | \mathbf{x}_{1:T})}[ \log q(\mathbf{s}_t | \mathbf{s}_{<t},  f_{\mathbf{w}}(\mathbf{x}_t))- \log p(\mathbf{s}_t | \mathbf{s}_{<t}, \mathbf{z}_{t-1})]\\&
= \int  \prod_{\tilde{t}=1}^T q(\mathbf{z}_{\tilde{t}} | \mathbf{s}_{\tilde{t}}, \mathbf{z}_{{\tilde{t}}-1}, f_{\mathbf{w}}(\mathbf{x}_{\tilde{t}}))q(\mathbf{s}_{\tilde{t}} | \mathbf{s}_{<{\tilde{t}}}, \mathbf{z}_{{\tilde{t}}-1}, f_{\mathbf{w}}(\mathbf{x}_{\tilde{t}})) \\&  [ \log q(\mathbf{s}_t | \mathbf{s}_{<t}, f_{\mathbf{w}}(\mathbf{x}_t))- \log p(\mathbf{s}_t | \mathbf{s}_{<t}, \mathbf{z}_{t-1})] \\&
= \int q(\mathbf{z}_{{t-1}} | \mathbf{s}_{{t-1}}, \mathbf{z}_{{{t}}-2}, f_{\mathbf{w}}(\mathbf{x}_{{t-1}}))\prod_{\tilde{t}=1}^t q(\mathbf{s}_{\tilde{t}} | \mathbf{s}_{<{\tilde{t}}}, \mathbf{z}_{{\tilde{t}}-1}, f_{\mathbf{w}}(\mathbf{x}_{\tilde{t}})) 
\\& [ \log q(\mathbf{s}_t | \mathbf{s}_{<t}, f_{\mathbf{w}}(\mathbf{x}_t))- \log p(\mathbf{s}_t | \mathbf{s}_{<t}, \mathbf{z}_{t-1})] \\&
= \mathbb{E}_{q(\mathbf{s}_1| f_{\mathbf{w}}(\mathbf{x}_1))}[...\mathbb{E}_{q(\mathbf{s}_{{t}} | \mathbf{s}_{<{{t}}}, \mathbf{z}_{{{t}}-1}, f_{\mathbf{w}}(\mathbf{x}_{{t}}))}[\mathbb{E}_{q(\mathbf{z}_{{t-1}} | \mathbf{s}_{{t-1}}, \mathbf{z}_{{{t}}-2}, f_{\mathbf{w}}(\mathbf{x}_{{t-1}}))}\\& [\textrm{KL}(q(\mathbf{s}_t | \mathbf{s}_{<t},f_{\mathbf{w}}(\mathbf{x}_t))|| p(\mathbf{s}_t | \mathbf{s}_{<t}, \mathbf{z}_{t-1}))]]]
    \end{aligned}
\end{equation}
The full ELBO for a single sequence is:
\begin{equation}
    \begin{aligned}
    \mathcal{L}_{ELBO} &= \sum^T_{t=1}  \mathbb{E}_{q(\mathbf{z}_t | \mathbf{s}_t, \mathbf{z}_{t-1}, f_{\mathbf{w}}(\mathbf{x}_t)) q(\mathbf{s}_t | \mathbf{s}_{<t}, f_{\mathbf{w}}(\mathbf{x}_t))}[\log p(\mathbf{x}_t | \mathbf{s}_t, \mathbf{z}_t)]\\
    &- \mathbb{E}_{ q(\mathbf{s}_{t} | \mathbf{s}_{<{t}}, \mathbf{z}_{{t}-1}, f_{\mathbf{w}}(\mathbf{x}_{t-1}))}[\mathbb{E}_{q(\mathbf{z}_{{t-1}} | \mathbf{s}_{{t-1}}, \mathbf{z}_{{{t-2}}}, f_{\mathbf{w}}(\mathbf{x}_{{t-1}}))} \\ & [\textrm{KL}(q(\mathbf{z}_t | \mathbf{s}_t, \mathbf{z}_{t-1}, f_{\mathbf{w}}(\mathbf{x}_t))|| p(\mathbf{z}_t|\mathbf{s}_t,\mathbf{z}_{t-1}))]]\\
    &- \mathbb{E}_{q(\mathbf{s}_1| f_{\mathbf{w}}(\mathbf{x}_1))}[...\mathbb{E}_{q(\mathbf{s}_{{t}} | \mathbf{s}_{<{{t}}}, \mathbf{z}_{{{t}}-1}, f_{\mathbf{w}}(\mathbf{x}_{{t}}))}[\mathbb{E}_{q(\mathbf{z}_{{t-1}} | \mathbf{s}_{{t-1}}, \mathbf{z}_{{{t}}-2}, f_{\mathbf{w}}(\mathbf{x}_{{t-1}}))}\\ & [\textrm{KL}(q(\mathbf{s}_t | \mathbf{s}_{<t}, \mathbf{z}_{t-1}, f_{\mathbf{w}}(\mathbf{x}_t))|| p(\mathbf{s}_t | \mathbf{s}_{<t}, \mathbf{z}_{t-1}))]]]
    \end{aligned}
\end{equation}